\documentclass[conference]{IEEEtran}
\IEEEoverridecommandlockouts
\usepackage{cite}
\usepackage{amsmath,amssymb,amsfonts}
\usepackage{algorithmic}
\usepackage{graphicx}
\usepackage{textcomp}
\usepackage{xcolor}
\usepackage{booktabs} 
\usepackage{multirow}
\usepackage{float}
\usepackage{hyperref}
\usepackage{listings}
\usepackage{tikz}
\usepackage{pgfplots}
\usepackage{enumitem}

\pgfplotsset{compat=1.18}
\usetikzlibrary{shapes.geometric, arrows.meta, positioning, calc, backgrounds, fit, patterns}

\setlist[itemize]{topsep=3pt, itemsep=2pt, partopsep=0pt, parsep=0pt}
\setlist[enumerate]{topsep=3pt, itemsep=2pt, partopsep=0pt, parsep=0pt}

\def\BibTeX{{\rm B\kern-.05em{\sc i\kern-.025em b}\kern-.08em
    T\kern-.1667em\lower.7ex\hbox{E}\kern-.125emX}}

\begin{document}

\title{Edge-Native Multimodal RAG: Hardware-Aware Quantization and Architectural Optimization for Consumer GPUs
\thanks{\copyright~2026 IEEE. Personal use of this material is permitted. Permission from IEEE must be obtained for all other uses, in any current or future media, including reprinting/republishing this material for advertising or promotional purposes, creating new collective works, for resale or redistribution to servers or lists, or reuse of any copyrighted component of this work in other works.}
}
\author{\IEEEauthorblockN{Zahraa Selim, Menna Hamed, Wesam Ahmed, Sohyla Said, Sara Basheer, and Rami Zewail}
\IEEEauthorblockA{\textit{Computer Science and Engineering Department} \\
\textit{Egypt-Japan University of Science and Technology (E-JUST)}\\
Alexandria, Egypt \\
\{zahraa.selim, menna.hamed, wesam.elraghy, sohyla.abdelwahed, sara.basheer, rami.zewail\}@ejust.edu.eg}
}

\maketitle

\begin{abstract}
Deployment of retrieval-augmented generation (RAG) for Edge-AI is constrained by memory and bandwidth limitations. A standard 7B parameter model consumes 14 GB Video Random Access Memory (VRAM) in half-precision (FP16), precluding co-location of vector indices and Key-Value (KV) caches. This paper presents an engineering integration and empirical evaluation of a multimodal RAG framework optimized for consumer-grade Turing hardware (e.g., NVIDIA Tesla T4). Our system-level analysis identifies a critical trade-off: while generative pre-trained transformer quantization (GPTQ) offers superior throughput ($51.6$ tok/s), it exhibits significant context instability at 2048+ tokens on Turing architectures. Conversely, NormalFloat 4-bit (NF4) maintains robust long-context performance with viable latency ($81$ ms/tok). By coupling NF4 with a decoupled Florence-2 vision integration, we achieve visual reasoning with only 430 MB VRAM overhead. Furthermore, empirical observations of prompt-induced search space pruning show that structured prompting reduces latency by 47\% without altering computational intensity (1.88 GFLOPs/token). The integrated architecture maintains an 8.37 GB footprint, establishing a viable RAG pipeline for 12--16 GB GPUs. Finally, structured few-shot prompting effectively compensates for quantization-induced reasoning loss, restoring mathematical accuracy from 35.0\% to 92.0\% across 200 trials.
\end{abstract}


\section{Introduction}
The proliferation of open-weight large language models (LLMs) has catalyzed a paradigm shift in the deployment of artificial intelligence. Compact architectures, such as Mistral-7B \cite{jiang2023mistral}, demonstrate that advanced capabilities are no longer the exclusive preserve of hyperscale cloud infrastructure. However, migrating these capabilities to consumer-grade hardware—typified by the NVIDIA T4 or RTX 3060—exposes  critical resource bottlenecks.

A standard seven-billion parameter model utilizing FP16 precision imposes a static memory footprint of approximately 14 GB. This baseline consumption virtually exhausts the memory budget of a 16-GB device, leaving negligible headroom for inference. This resource contention is exacerbated in retrieval-augmented generation (RAG) pipelines \cite{lewis2020rag}, where the system must simultaneously accommodate embedding models ($\approx$300 MB), vector indices ($\approx$200 MB), and context-dependent Key-Value (KV) caches. A naive implementation inevitably exceeds the physical VRAM limits of ubiquitous 12--16-GB GPUs, often demanding an aggregate footprint of 17--19 GB \cite{kim2023squeezellm}.

\subsection{Motivation and Research Gaps}
This study is predicated on addressing three persistent deployment lacunae through systems engineering and empirical validation:
\begin{itemize}
    \item \textbf{Architectural Agnosticism in Quantization:} Techniques optimized for specific kernels often incur uncharacterized stability penalties when deployed on older hardware, such as NVIDIA Turing GPUs.
    \item \textbf{Monolithic VRAM Bottlenecks:} End-to-end multimodal LLMs drastically exceed consumer VRAM budgets, precluding their use in local, disconnected RAG environments.
    \item \textbf{Silent Generative Failures:} Prevalent evaluation relies heavily on perplexity, obfuscating how aggressive bit-width reduction specifically degrades multi-step reasoning and faithfulness in retrieval tasks.
\end{itemize}

\subsection{Contributions}
To bridge these gaps, we propose a tripartite system integration framework. Our primary contributions correlate directly to the identified gaps:
\begin{enumerate}
    \item \textbf{Hardware-Aware Quantization Benchmarking (Addressing Gap 1):} We isolate a critical stability bottleneck on the NVIDIA Turing architecture. While GPTQ provides the fastest inference ($19.36$ ms/tok), it exhibits high variance and context collapse. NF4 serves as the optimal middle ground, maintaining viable edge performance ($81$ ms/tok) with superior stability.
    \item \textbf{Decoupled Multimodal Architecture (Addressing Gap 2):} We integrate a decoupled visual pipeline using Florence-2 Base, achieving visual QA capabilities comparable to larger models with only a negligible 430 MB VRAM overhead.
    \item \textbf{Reasoning Restoration and Prompt Efficiency (Addressing Gap 3):} We demonstrate that while quantization degrades procedural reasoning, structured few-shot prompting compensates for this loss, recovering mathematical accuracy from 35.0\% to 92.0\% across 200 trials. Concurrently, we document a prompt-induced efficiency gain, reducing generation latency by 47\% via search-space pruning.
\end{enumerate}

\section{Related Work}
Deploying LLMs in resource-constrained edge environments necessitates a rigorous reduction in computational complexity \cite{zhu2023survey}. Techniques such as Quantization,Prunning, and Knowledge Distillation  have emerged as a highly viable strategy for immediate memory relief without necessitating extensive architectural retraining. 

\subsection{Quantization Strategies}
Quantization fundamentally alters the arithmetic intensity of inference by lowering the precision of parameters. Weight-only strategies compress static model parameters while retaining high-precision activations. A prominent example is GPTQ \cite{frantar2023gptq}, which minimizes reconstruction error by exploiting second-order Hessian information. Mixed-precision decomposition techniques have emerged as a solution to address precision loss, such as LLM.int8() \cite{dettmers2022llmint8} and SpQR \cite{dettmers2024spqr}. 

Standard weight-only methods often falter when confronted with activation outliers. Activation-aware strategies postulate that preserving the fidelity of these outliers is paramount, as seen in AWQ \cite{lin2023awq}. Complementary approaches, such as SmoothQuant \cite{xiao2023smoothquant}, employ mathematical transformations to smooth the activation landscape. In RAG workflows, the linear growth of the KV cache constitutes a formidable memory bottleneck, prompting specialized algorithms such as KVQuant \cite{lee2024kvquant} and KIVI \cite{liu2024kivi}, though integrating these natively with consumer hardware APIs remains complex.

\subsection{Pruning and Knowledge Distillation}
Pruning methodologies allow for the excision of redundant weights. Unstructured pruning induces sparsity at the scalar level (e.g., SparseGPT \cite{frantar2023sparsegpt}), while structured pruning removes coherent architectural blocks (e.g., LLM-Pruner \cite{ma2023llmpruner}). Knowledge distillation functions to compress the semantic density of a "teacher" model. Techniques such as Fine-tune-CoT \cite{lobo2024impact} and Chain of Preference Optimization \cite{zhang2024cpo} are designed to explicitly optimize the student's ability to emulate complex reasoning paths.

\subsection{Resource-Constrained Edge Learning}
Beyond LLM-specific compression, the broader paradigm of resource-constrained learning has seen significant exploration across various edge applications. Building on our previous work, the synergy of sparsity and quantization has been extensively validated for enabling lightweight edge deployment across various domains \cite{aboutahoun2023lightweight,herman2023lightweight}. Furthermore, our prior studies have demonstrated the considerable potential of few-shot learning in reducing computational dependencies for Edge-AI in both medical \cite{aboutahoun2022potentials} and industrial contexts \cite{zewail2026resource}. The current work adapts these foundational principles from our group—specifically few-shot prompting and post-training quantization—to system-level RAG integration.

\section{System Architecture and Methodology}
To rigorously evaluate the interplay between hardware constraints and generation fidelity, this study adopted a bifurcated empirical methodology, decoupling the arithmetic impact of quantization on latency from its downstream semantic consequences.

\subsection{Quantization Benchmarking Study}
In the initial phase, we employed the foundational \textbf{Mistral-7B-v0.1} model \cite{jiang2023mistral}. This base model was intentionally selected for the benchmarking phase to evaluate the raw weight distribution entropy without the bias of instruction-tuning interventions. Benchmarking was executed exclusively on an \textbf{NVIDIA Tesla T4 (16 GB GDDR6)} to emulate a constrained edge environment. 

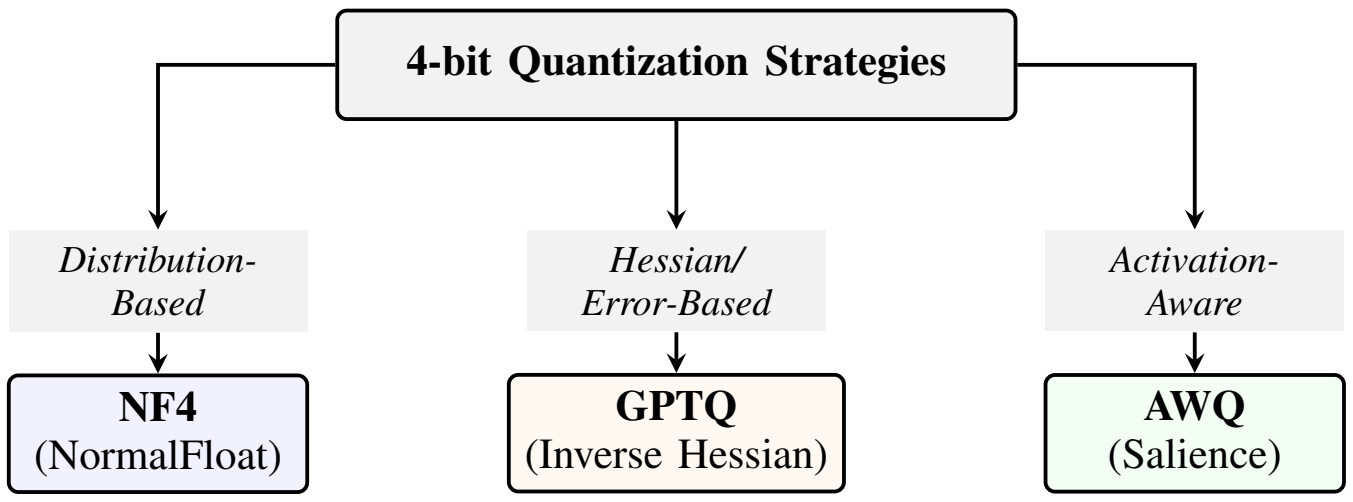
\begin{figure}[h]
\centering
\resizebox{\columnwidth}{!}{%
\begin{tikzpicture}[
    node distance=0.6cm and 0.4cm,
    box/.style={rectangle, draw=black, thick, rounded corners=2pt, minimum width=2.2cm, minimum height=0.8cm, align=center, font=\small},
    cat/.style={rectangle, draw=none, fill=gray!10, minimum width=2.2cm, minimum height=0.6cm, align=center, font=\footnotesize\itshape},
    arrow/.style={thick, ->, >=stealth}
]
\node (root) [box, minimum width=5cm, fill=black!5] {\textbf{4-bit Quantization Strategies}};

\node (dist) [cat, below left=0.8cm and 0.2cm of root] {Distribution-\\Based};
\node (hess) [cat, below=0.8cm of root] {Hessian/\\Error-Based};
\node (act) [cat, below right=0.8cm and 0.2cm of root] {Activation-\\Aware};

\node (nf4) [box, below=0.3cm of dist, fill=blue!5] {\textbf{NF4}\\(NormalFloat)};
\node (gptq) [box, below=0.3cm of hess, fill=orange!5] {\textbf{GPTQ}\\(Inverse Hessian)};
\node (awq) [box, below=0.3cm of act, fill=green!5] {\textbf{AWQ}\\(Salience)};

\draw [arrow] (root) -| (dist);
\draw [arrow] (root) -- (hess);
\draw [arrow] (root) -| (act);

\draw [arrow] (dist) -- (nf4);
\draw [arrow] (hess) -- (gptq);
\draw [arrow] (act) -- (awq);
\end{tikzpicture}%
}
\caption{Classification of evaluated quantization techniques.}
\label{fig:taxonomy}
\end{figure}

We evaluated three PTQ paradigms (Fig. \ref{fig:taxonomy}):
\begin{itemize}
    \item \textbf{NF4:} Employs nested quantile quantization optimized for zero-mean normal distributions.
    \item \textbf{GPTQ:} A layer-wise reconstruction algorithm ($g=128$, $\lambda=0.01$) calibrated on WikiText-2 \cite{merity2016wikitext}.
    \item \textbf{AWQ:} An activation-aware approach protecting the top 1\% of weights based on activation magnitude.
\end{itemize}

\subsection{Multimodal RAG Architecture}
\begin{figure*}[t]
\centering
\resizebox{\textwidth}{!}{%
\begin{tikzpicture}[
    node distance=1.5cm,
    process/.style={rectangle, draw=black!60, fill=white, thick, minimum width=2.5cm, minimum height=1.2cm, align=center, font=\small, rounded corners=3pt},
    storage/.style={cylinder, draw=black!60, fill=gray!10, thick, aspect=0.25, minimum height=1.5cm, minimum width=2.0cm, shape border rotate=90, align=center, font=\small},
    arrow/.style={thick, ->, >=stealth, color=black!80},
]
\node (docs) [process] {Source Docs\\(PDF)};
\node (chunk) [process, right=1.0cm of docs] {Semantic\\Chunking\\(512 tokens)};
\node (embed_idx) [process, right=1.0cm of chunk] {Embedding\\Model\\(MiniLM-L6)};
\node (db) [storage, right=1.0cm of embed_idx] {Vector\\Index\\(ChromaDB)};
\node (retrieve) [process, below=1.8cm of db] {Retrieval\\(Top-K)};
\node (embed_qry) [process, left=1.0cm of retrieve] {Query\\Embedding};
\node (query) [process, left=1.0cm of embed_qry] {User\\Query};

\node (llm) [process, right=1.0cm of retrieve, fill=blue!5, thick] {Quantized LLM\\(Mistral-7B)};
\node (output) [process, right=1.0cm of llm] {Generated\\Answer};

\draw [arrow] (docs) -- (chunk);
\draw [arrow] (chunk) -- (embed_idx);
\draw [arrow] (embed_idx) -- (db);

\draw [arrow] (query) -- (embed_qry);
\draw [arrow] (embed_qry) -- (retrieve);
\draw [arrow] (db) -- (retrieve) node[midway, right, font=\footnotesize] {Search};

\draw [arrow] (retrieve) -- (llm) node[midway, above, font=\footnotesize] {Context};
\draw [arrow] (llm) -- (output);
\draw [arrow] (query) to[out=30, in=150] node[midway, above, font=\footnotesize, text=black] {Prompt Text} (llm);

\node [above=0.2cm of docs, font=\bfseries\small, color=gray] {Phase 1: Indexing};
\node [above=0.2cm of query, font=\bfseries\small, color=gray] {Phase 2: Inference};
\end{tikzpicture}%
}
\caption{The Modular RAG Pipeline Architecture. Indexing (Phase 1) is decoupled from inference (Phase 2).}
\label{fig:rag_pipeline}
\end{figure*}
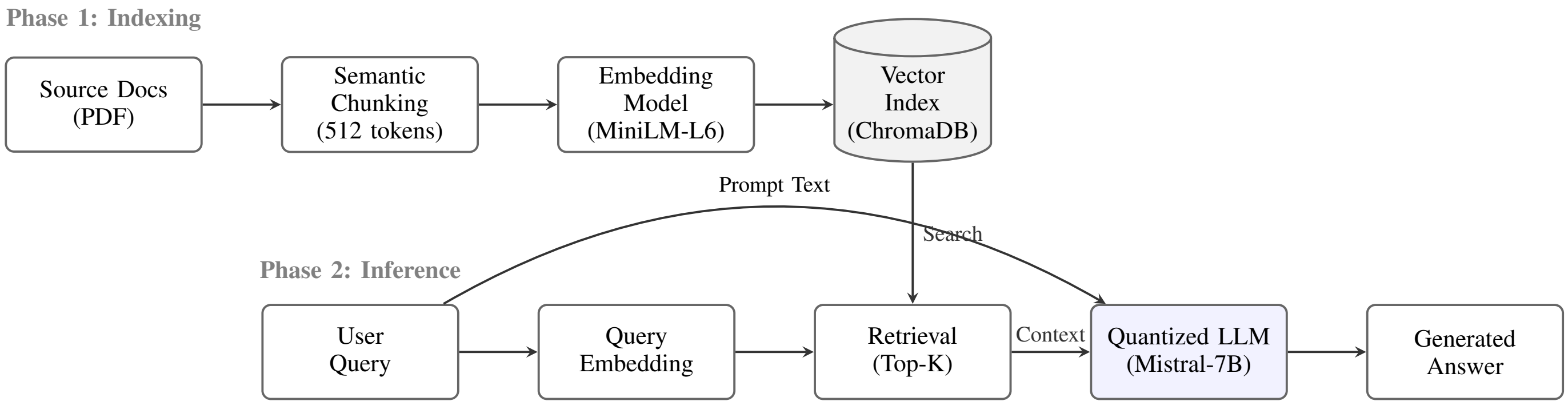

Leveraging the stability findings from Phase 1, we architected the final system to prioritize stability over raw throughput. The production pipeline employs the \textbf{Mistral-7B-Instruct-v0.2} checkpoint to natively support Sliding Window Attention (SWA) for 32k context handling and to ensure strict adherence to retrieval constraints. The system orchestrates a modular pipeline operating strictly within a 16 GB envelope (Fig. \ref{fig:rag_pipeline}).

\subsubsection{Adaptive Ingestion and Dual-Path Retrieval}
Documents undergo heuristic classification to route parsing tasks between fast text extraction (\texttt{PyMuPDF}) and structure-preserving Optical Character Recognition (OCR) (\texttt{Marker}/\texttt{Surya}). The integrity of the ingested text is verified via an empirical coherence metric, $Q_{\text{extract}}$:
\begin{equation}
    Q_{\text{extract}} = \frac{C_{\text{valid}}}{C_{\text{total}}} \times \frac{S_{\text{count}}}{S_{\text{exp}}}
\end{equation}
This serves as a heuristic baseline to trigger OCR retries if the score falls below $\tau=0.5$. Content is chunked dynamically (256--768 tokens) to prevent breaking semantic structures like tables.

Retrieval utilizes a dual-path mechanism combining dense vectors (\textit{BGE-small-en-v1.5} \cite{xiao2023cpack}) and sparse lexical representations (BM25 \cite{robertson2009bm25}). The disparate scores are fused using Reciprocal Rank Fusion (RRF) \cite{cormack2009rrf}:
\begin{equation}
    \text{RRF}(d) = \sum_{r \in \{\text{dense}, \text{sparse}\}} \frac{1}{60 + \text{rank}_r(d)}
\end{equation}

\subsubsection{Decoupled Multimodal Integration}
\begin{figure}[htbp]
\centering
\resizebox{0.95\columnwidth}{!}{%
\begin{tikzpicture}[
    node distance=1.1cm,
    box/.style={rectangle, draw=black!70, fill=white, thick, minimum width=2.4cm, minimum height=0.8cm, align=center, font=\small, rounded corners=2pt},
    arrow/.style={thick, ->, >=stealth, color=black!70}
]
\node (img) [box, fill=gray!5] {Document Image};
\node (florence) [box, right=0.8cm of img, fill=green!5, draw=green!60!black] {Florence-2 Base};
\node (fuse) [box, below=0.8cm of florence, fill=orange!5] {Multimodal Fusion};
\node (text) [box, left=0.8cm of fuse] {OCR Output};
\node (llm) [box, right=0.8cm of fuse, fill=blue!5, draw=blue!60!black, ultra thick] {Quantized LLM\\(Mistral NF4)};

\draw [arrow] (img) -- (florence);
\draw [arrow] (florence) -- (fuse) node[midway, right, font=\tiny] {Text Tokens};
\draw [arrow] (text) -- (fuse);
\draw [arrow] (fuse) -- (llm);
\end{tikzpicture}
}
\caption{Decoupled Vision-Language Architecture.}
\label{fig:vision_arch}
\end{figure}
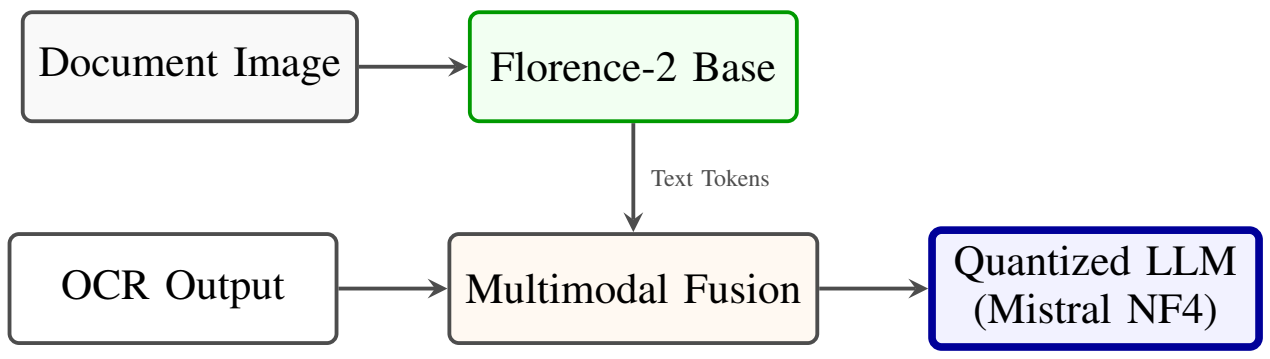

To endow the system with visual reasoning without the prohibitive memory demands of monolithic VLMs, we implemented a decoupled topology utilizing \textbf{Microsoft Florence-2 Base} (0.23B parameters) \cite{xiao2023florence} (Fig. \ref{fig:vision_arch}). 

Images are processed via chained prompts (\texttt{<OD>}, \texttt{<MORE\_DETAILED\_CAPTION>}) to extract spatial grounding and narrative semantics. The serialized outputs are injected into a standardized textual template, effectively "translating" visual features into text. This incurs a negligible VRAM overhead of approximately \textbf{430 MB}.

\subsubsection{Structured Prompting and Fidelity Validation}
To counteract reasoning degradation inherent in 4-bit models, we deploy structured few-shot prompting, enforcing intent-driven output topologies. The system employs a post-generation heuristic filter ($Q_{\text{valid}}$), assessing candidates against an empirical threshold of $\tau=0.6$:
\begin{equation}
    Q_{\text{valid}} = \sum_{d \in D} w_d \cdot S_d
\end{equation}
While these dimensions ($w_d$) were heuristically selected based on empirical observations in our target domain (e.g., Groundedness $w=0.5$), they provide a functional baseline structural filter.

\section{Experiments and Results}

\subsection{Quantization Benchmarking Analysis}
Latency and throughput were evaluated on a Tesla T4. As shown in Table \ref{tab:inference_performance}, GPTQ provides the highest theoretical throughput ($51.65$ tok/s) and the best Time to First Token (TTFT). However, empirical evaluation revealed severe instability and F1-score collapse at context lengths exceeding 2048 tokens on this architecture. Consequently, NF4, while slower ($12.29$ tok/s), was selected for the RAG pipeline to ensure deterministic system reliability. 

\begin{table}[h]
\centering
\caption{Inference Performance and Hardware Compatibility (Tesla T4)}
\label{tab:inference_performance}
\footnotesize
\begin{tabular}{lccc}
\toprule
\textbf{Model} & \textbf{Latency (ms/tok)} & \textbf{Throughput (tok/s)} & \textbf{TTFT (ms)} \\
\midrule
FP16 Baseline & $63.45 \pm 0.20$ & $15.76$ & $71.19$ \\
\textbf{NF4 (Ours)} & \textbf{81.37} $\pm$ \textbf{0.95} & \textbf{12.29} & \textbf{178.09} \\
AWQ & $102.96 \pm 1.13$ & $9.71$ & $114.31$ \\
GPTQ & $19.36 \pm 0.14$ & $51.65$ & $43.02$ \\
\bottomrule
\end{tabular}
\end{table}

\subsection{System-Level Memory Feasibility}
The breakdown of the aggregate memory footprint (Table \ref{tab:memory}) confirms the feasibility of edge deployment. The complete multimodal pipeline incurs a peak VRAM load of 8.37 GB on the Tesla T4. This footprint theoretically accommodates the tighter 12-GB envelope of consumer GPUs like the RTX 3060.

\begin{table}[htbp]
\caption{Complete System Memory Budget (Tesla T4)}
\label{tab:memory}
\centering
\footnotesize
\begin{tabular}{lcc}
\toprule
\textbf{Component} & \textbf{Memory (GB)} & \textbf{Share (\%)} \\
\midrule
Quantized LLM (Mistral-7B) & 3.70 & 44.2\% \\
CUDA Context \& Kernels & 2.84 & 33.9\% \\
Vision Agent (Florence-2) & 0.43 & 5.1\% \\
KV Cache (4k Context) & 0.40 & 4.8\% \\
Embedding Model (BGE) & 0.30 & 3.6\% \\
Vector Index (FAISS) & 0.20 & 2.4\% \\
System Overhead & 0.50 & 6.0\% \\
\midrule
\textbf{Total Footprint} & \textbf{8.37} & \textbf{100.0\%} \\
\bottomrule
\end{tabular}
\end{table}

\subsection{Retrieval Fidelity and Answer Quality}
The RRF hybrid retrieval engine achieved a Recall@5 of 0.850 on SQUADv2 \cite{rajpurkar2018squad} (Table \ref{tab:retrieval_quality}). Generative fidelity data (Table \ref{tab:rag_quality}) reveal that the NF4 quantized Instruct model incurs only a slight degradation compared to the FP16 baseline (F1 $0.157$ vs. $0.164$). 

\begin{table}[h]
\centering
\caption{Retrieval Quality Metrics (SQUADv2)}
\label{tab:retrieval_quality}
\footnotesize
\begin{tabular}{lcc}
\toprule
\textbf{Metric} & \textbf{Score} & \textbf{Implication} \\
\midrule
Recall@5 (Hybrid) & 0.850 & RRF Fusion of Dense + Sparse \\
Recall@5 (Dense) & 0.700 & Semantic Embedding (BGE) \\
Recall@5 (Sparse) & 0.650 & Keyword Matching (BM25) \\
\bottomrule
\end{tabular}
\end{table}

\begin{table}[h]
\centering
\caption{RAG Answer Quality (Generative Fidelity)}
\label{tab:rag_quality}
\footnotesize
\begin{tabular}{lcc}
\toprule
\textbf{Model} & \textbf{F1 Score} & \textbf{Faithfulness} \\
\midrule
FP16 Baseline & 0.164 & 0.618 \\
\textbf{NF4 (Ours)} & \textbf{0.157} & \textbf{0.583} \\
AWQ & 0.168 & 0.604 \\
\bottomrule
\end{tabular}
\end{table}

\subsection{Reasoning Restoration and Prompt Efficiency}
We investigated the impact of structured prompting on reasoning accuracy and computational cost (Table \ref{tab:prompting_impact}). Computational intensity remained invariant at 1.88 GFLOPs/tok, yet mean total generation time dropped by 47\% (from 20.06 s to 10.64 s). We attribute this to prompt-induced search-space pruning; forcing explicit structural constraints eliminates verbose hallucination loops. Crucially, this structured few-shot formatting successfully compensates for the reasoning loss observed in zero-shot NF4 models, restoring accuracy on empirical mathematical tasks from 35.0\% to 92.0\% over 200 trials.

\begin{table*}[t]
\centering
\caption{Impact of Structured Prompting on Efficiency \& Reasoning}
\label{tab:prompting_impact}
\footnotesize
\begin{tabular}{lcccc}
\toprule
\textbf{Metric Domain} & \textbf{Metric Name} & \textbf{Math Baseline (Zero-Shot)} & \textbf{Structured (Few-Shot)} & \textbf{Delta / Improvement} \\
\midrule
\multirow{3}{*}{\textbf{Reasoning Performance}} 
  & Math Accuracy ($N=50$) & 35.0\% (Mean) & 92.0\% (Mean) & +57.0 pp \\
  & Reasoning Consistency & $\pm$1.0\% (18 vs 17) & 0.0\% (46 vs 46) & -1.0 pp \\
  & Reasoning Faithfulness & 0.216 & 0.915 & +323\% \\
\midrule
\multirow{3}{*}{\textbf{System Efficiency}} 
  & Perplexity (WikiText-2) & 13.02 & 6.67 & -48.8\% \\
  & Computational Cost & 1.88 GFLOPs/tok & 1.88 GFLOPs/tok & 0.0\% \\
  & Avg. Generation Time & 20.06 s & 10.64 s & -47.0\% \\
\bottomrule
\end{tabular}
\end{table*}

\subsection{Multimodal Integration and Comparative Efficacy}
To validate the decoupled system, we evaluated a custom dataset of 50 tri-modal QA dyads. As shown in Table \ref{tab:vision_perf}, the integration of Florence-2 Base yielded a 192.7\% accuracy uplift relative to the text-only baseline. While native monolithic VLMs offer higher theoretical ceilings, their massive footprint excludes them from this 8GB hardware budget. The decoupled approach achieves functional parity (0.249 vs 0.280 score) with the 1.86B-parameter Moondream2 model \cite{moondream2024} while consuming approximately 50\% less VRAM.

\begin{table*}[t]
\centering
\caption{Comparative Evaluation of Vision Encoders (T4 GPU)}
\label{tab:vision_perf}
\footnotesize
\begin{tabular}{lcccccc}
\toprule
\textbf{Model Architecture} & \textbf{Parameters} & \textbf{Peak VRAM} & \textbf{Vision-Only Score} & \textbf{Overall RAG Score} & \textbf{Relative Gain} & \textbf{Deployment Status} \\
\midrule
No Vision (Baseline) & -- & 2.47 GB & 0.090 & 0.235 & -- & Feasible (Text-Only) \\
\textbf{Florence-2 Base} \cite{xiao2023florence} & \textbf{0.23B} & \textbf{2.90 GB} & \textbf{0.265} & \textbf{0.249} & \textbf{+192.7\%} & \textbf{Selected (Ours)} \\
Florence-2 Large \cite{xiao2023florence} & 0.77B & 3.92 GB & 0.256 & 0.263 & +183.1\% & Feasible \\
Moondream2 \cite{moondream2024} & 1.86B & 6.07 GB & 0.256 & 0.280 & +183.7\% & Constrained \\
ViT-GPT2 \cite{nlpconnect2024} & 0.16B & 2.94 GB & 0.251 & 0.210 & +177.6\% & Feasible \\
\bottomrule
\end{tabular}
\end{table*}

\section{Discussion, Limitations, and Future Work}

\subsection{Architectural Bottlenecks}
A granular decomposition of latency exposes a critical trade-off. While GPTQ offers superior throughput, empirical benchmarking reveals a critical stability failure at extended context lengths on the Turing architecture. We explicitly prioritize NF4 to guarantee inferential robustness, trading peak throughput for system reliability. Furthermore, we demonstrate that structural constraints effectively compensate for the quantization-induced loss of parametric confidence without requiring weight-adjustment methods.

\subsection{Limitations }
While this study establishes a functional engineering baseline, several limitations warrant future exploration. First, the root-cause of GPTQ context collapse (e.g., quantization noise accumulation versus ExLlamaV2 kernel bugs) requires low-level profiling outside the scope of this system integration. Second, to maintain simplicity on older consumer APIs, the KV cache remained in uncompressed FP16; integrating explicit KV cache quantization (e.g., KIVI) or newer formats like Q4\_K\_M remains an open avenue for further memory reduction. Third, mathematical evaluation relied on a domain-specific dataset rather than standard benchmarks like GSM8K to target distinct reasoning failures. Finally, while static estimations suggest viability on the RTX 3060, comprehensive end-to-end testing across Ada and Ampere architectures is required to validate generalizability.

\section{Conclusion}
This study provides an empirical evaluation of retrieval-augmented generation on memory-constrained edge hardware. By articulating a systems-engineering framework spanning NF4 arithmetic quantization, decoupled visual integration, and prompt-driven search-space pruning, we establish that functional RAG is attainable on ubiquitous 12--16 GB GPUs within a sustainable \textbf{8.37 GB operational envelope}. We identify  a severe context degradation with GPTQ on Turing architectures, validating NF4 as a robust alternative. Most significantly, we demonstrate that "reasoning degradation" is reversible via structured few-shot prompting, restoring accuracy across 200 mathematical trials while simultaneously accelerating generation latency by 47\%.
As a future direction, the proposed framework opens the door for exploring the potential of lightweight LLMs in heavy industries such as the oil and gas sector, where an edge-native RAG pipeline can be integrated with autonomous downhole systems  to turn complex, unstructured sensor data into actionable insights with minimal cloud reliance \cite{alamin2025llm} .

\section*{Acknowledgment}
The author(s) acknowledge the use of Paperpal for grammar correction and stylistic refinement during the preparation of this manuscript. The author(s) have reviewed and edited the content and assume full responsibility for the final publication.

\bibliographystyle{IEEEtran}
\bibliography{wip_journal_version_references}

\end{document}